\documentclass[conference]{IEEEtran}
\usepackage{blindtext, graphicx}
\usepackage[utf8]{inputenc}
\usepackage[english]{babel}
\usepackage{listings}
\usepackage{amssymb}
\usepackage[breaklinks=true]{hyperref}
\usepackage[linesnumbered,ruled,vlined]{algorithm2e}
\usepackage{amsmath}
\usepackage{hyperref}
\usepackage{booktabs}

\begin{document}

\title{{A new training approach for text classification in Mental Health: LatentGLoss}}

\author{
Korhan Sevinc$^{1}$\\ \\
\small{$^{1}$TOBB University of Economics and Technology}\\
\small{\texttt{ksevinc@etu.edu.tr}}
}
\author{\IEEEauthorblockN{Korhan Sevinc}
\IEEEauthorblockA{Department of Computer Engineering \\
TOBB University of Economics and Technology\\
Ankara, Turkey\\
ksevinc@etu.edu.tr}
}

\maketitle

\begin{abstract}
This study presents a multi-stage approach to mental health classification by leveraging traditional machine learning algorithms, deep learning architectures, and transformer-based models. A novel data set was curated and utilized to evaluate the performance of various methods, starting with conventional classifiers and advancing through neural networks. To broaden the architectural scope, recurrent neural networks (RNNs) such as LSTM and GRU were also evaluated to explore their effectiveness in modeling sequential patterns in the data. Subsequently, transformer models such as BERT were fine-tuned to assess the impact of contextual embeddings in this domain. Beyond these baseline evaluations, the core contribution of this study lies in a novel training strategy involving a dual-model architecture composed of a teacher and a student network. Unlike standard distillation techniques, this method does not rely on soft label transfer; instead, it facilitates information flow through both the teacher model’s output and its latent representations by modifying the loss function. The experimental results highlight the effectiveness of each modeling stage and demonstrate that the proposed loss function and teacher-student interaction significantly enhance the model's learning capacity in mental health prediction tasks.

\end{abstract}

\IEEEpeerreviewmaketitle

\section{Introduction}
Mental well-being promotion is central to the action plans of both the World Health Organization (WHO) for the years 2013–2020 \cite{WHO2013} and the European Union's European Pact on Mental Health and Well-being  \cite{EU2008}. The key breakthrough in addressing mental health issues lies in developing tools for early detection and preventive measures  \cite{Insel2006}. The WHO action plan highlights the importance of health policies and programs that not only address the needs of those affected by mental health disorders but also work to preserve mental well-being. These policies emphasize evidence-based, non-pharmacological interventions, focusing on early intervention and preventing unnecessary institutionalization and medicalization. Successful interventions are particularly dependent on how frequently the therapy can be accessed  \cite{Hansen2002}. In this context, automated systems have significant advantages over traditional therapies due to their ability to provide continuous support at minimal additional cost. Consequently, health assistants capable of delivering therapeutic interventions have gained considerable attention in recent years  \cite{Bickmore2005} \cite{6}. However, these systems are predominantly based on manually designed rules. In contrast, research in statistical methods for conversational systems has largely been limited to narrow-domain information-seeking dialogues  \cite{7} \cite{8} \cite{9} \cite{10}.

In our study, we aim to extract insights from human conversations, texts, and writings to address mental health issues. For this purpose, a custom dataset has been collected, specifically designed to focus on mental health-related text data. Various machine learning (ML) and deep learning (DL) methods were tested on this dataset to evaluate their classification capabilities. Furthermore, transformer-based models, including BERT, were fine-tuned to capture deeper semantic relationships within the data.

Building on these efforts, a novel approach was introduced that leverages a dual-model architecture, consisting of a teacher model and a student model. This method utilizes both the predictions and latent vectors from the teacher model to transfer knowledge to the student model, aided by a modified loss function. This innovative strategy contributes significantly to improving the classification performance, addressing some of the key challenges identified in the mental health analysis field.

All experimental code, link to data, training configurations and model checkpoints, are shared and maintained publicly at:  
\href{https://github.com/korhansevinc/LatentG-Loss}{https://github.com/korhansevinc/LatentG-Loss}

\section{Related Work}

Mental health issues are a growing global concern, with increasing efforts to develop automated systems for early detection and intervention. One of the promising approaches in this domain involves the use of text classification techniques, which leverage natural language processing (NLP) models to analyze written content for signs of mental health disorders. Numerous studies have explored the potential of machine learning (ML) and deep learning (DL) methods for classifying text data, particularly in the context of mental health.

One significant study in this area is by Cohan et al. (2018), who used text classification methods to detect mental health conditions based on online forum posts. They employed traditional machine learning algorithms, such as Support Vector Machines (SVM), and demonstrated the efficacy of these models in identifying depression-related content from social media posts and online discussions. The authors highlighted the challenge of using unstructured textual data and discussed how linguistic features, such as sentiment, word choice, and the frequency of certain keywords, played a crucial role in the classification process  \cite{11}.

In a similar vein, Rea et al. (2019) explored the use of recurrent neural networks (RNNs) for mental health classification tasks. Their approach focused on using long short-term memory (LSTM) networks to capture the sequential nature of textual data, which is essential for understanding the context and progression of mental health issues. Their work showed that RNN-based architectures outperformed traditional ML models in identifying mental health disorders in text data  \cite{12}.

Another notable approach was proposed by Bhatia et al. (2020), who investigated the use of Transformer-based models, particularly BERT, for the classification of mental health-related text. By fine-tuning a pre-trained BERT model on mental health datasets, they were able to achieve state-of-the-art results in detecting various mental health conditions, including depression, anxiety, and stress, from text. Their results demonstrated the power of pre-trained language models in handling complex and nuanced language patterns present in mental health-related texts  \cite{13}.

Moreover, Dey and Desai (2022) used LSTM networks in combination with GloVe word embeddings to classify mental health issues from text data. Their approach focused on the use of rich semantic representations provided by GloVe embeddings, which significantly improved the performance of the LSTM model in detecting mental health issues. This study emphasized the importance of leveraging pre-trained word embeddings to capture the semantic meaning of text, highlighting the advantage of using these models in resource-limited settings  \cite{14}.

Furthermore, some studies have focused on developing multi-modal approaches to mental health analysis, combining text data with other sources such as images or audio. For instance, Kumar et al. (2021) presented a multi-modal deep learning framework that combined text and speech analysis to detect mental health disorders. Their model was able to capture both the linguistic and acoustic features from patient interactions, leading to improved classification performance  \cite{15}.

While these studies provide valuable insights into the use of text classification for mental health detection, it is important to note that even though these methods are well-regarded for mental health classification, achieving high accuracy and fast predictions based solely on constructed sentences or conversations remains challenging. In most cases, additional features or assumptions are required to improve the results. For example, linguistic and psychological features may need to be explicitly integrated into the models to achieve reliable predictions. In contrast, our approach aims to directly predict the mental health state of individuals from their constructed sentences, without relying on external features or assumptions, highlighting a more direct method for classification.

In addition to leveraging traditional ML and DL methods, a novel approach is proposed in this work, involving a dual architecture system. This system consists of a teacher model and a student model, where knowledge is transferred through the teacher's outputs and latent representations, guided by a modified loss function. This method aims to enhance model performance by leveraging the advantages of knowledge distillation while addressing the challenges posed by limited training data.

\section{Dataset}

The dataset used in this study is constructed by combining multiple publicly available mental health datasets, including \textit{Depression Reddit Cleaned}, \textit{Human Stress Prediction}, \textit{Predicting Anxiety in Mental Health}, \textit{Mental Health Bipolar}, \textit{Reddit Mental Health Data}, \textit{Students Anxiety and Depression}, \textit{Suicidal Mental Health}, \textit{Suicidal Tweet Detection}, and \textit{3k Conversations Dataset for Chatbot}. To build a unified and robust dataset, we applied various preprocessing and data augmentation techniques, merging all samples into a single consistent format.

The final version of the dataset consists of unique person IDs, their statements obtained from conversations, chat logs, and other text-based platforms, and the corresponding mental health labels. The classification includes the following seven mental health categories:

\begin{itemize}
    \item \textbf{Normal}
    \item \textbf{Depression}
    \item \textbf{Suicidal}
    \item \textbf{Anxiety}
    \item \textbf{Stress}
    \item \textbf{Bipolar}
    \item \textbf{Personal Disorder}
\end{itemize}

Each sample is composed of a sentence or a few sentences representing the mental state of an individual. Below are a few example excerpts from different categories:

\textbf{Anxiety:}
\begin{itemize}
    \item "All wrong, back off dear, forward doubt. Stay in a restless and restless place."
    \item "I'm restless and restless, it's been a month now, boy. What do you mean?"
    \item "I haven't slept well for 2 days, it's like I'm restless. Why huh :("
\end{itemize}

\textbf{Depression:}
\begin{itemize}
    \item "Please do not lecture me, I feel bad enough as it is. I was diagnosed with depression and anxiety approximately 14 years ago."
    \item "Every day it is just like a tug of war with myself."
    \item "Today is not a very good day, and I just want to share it with someone who cares."
\end{itemize}

\textbf{Suicidal:}
\begin{itemize}
    \item "I have been suicidal for what feels like no reason, but there is a reason..."
    \item "I keep trying to put it away or throw the knife away, but it is like an addiction."
    \item "I tried hanging myself when I was 16. The rope broke."
\end{itemize}

\textbf{Normal:}
\begin{itemize}
    \item "I haven't opened it for 2 days, it's all over, it's really late."
    \item "Only two days of fasting."
    \item "It's true."
\end{itemize}

\vspace{0.5cm}
\begin{figure}[ht]
\centering
\includegraphics[width=0.5\textwidth]{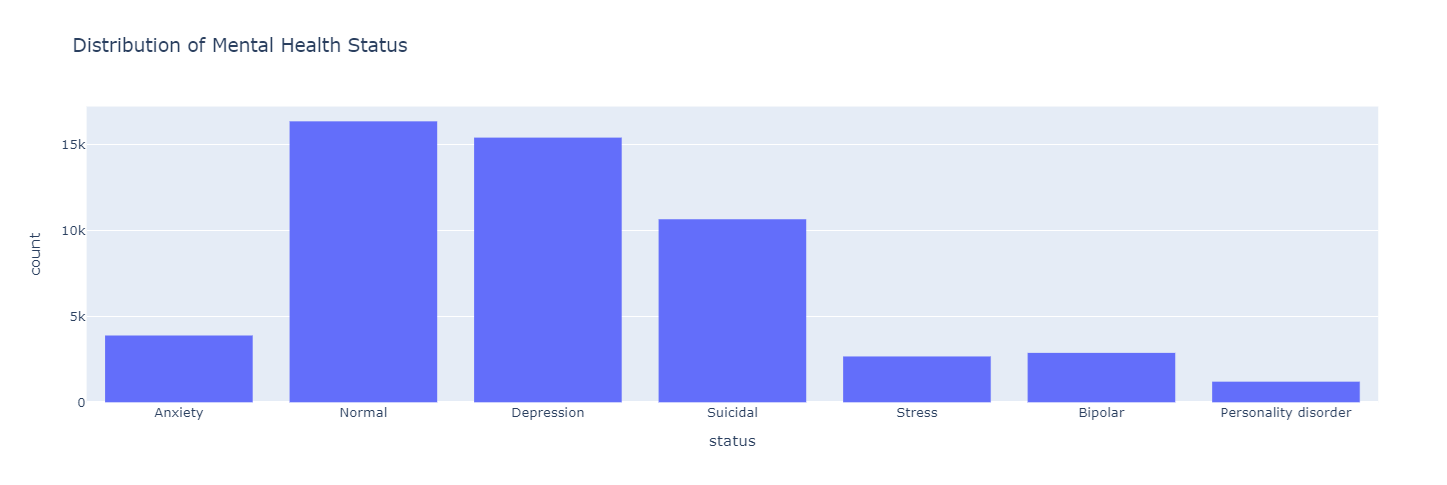}
\caption{Distribution of classes in the final dataset.}
\label{fig:class_distribution}
\end{figure}

As can be observed from Figure 1, the dataset exhibits a significant class imbalance problem, where certain mental health statuses such as "Normal" and "Depression" are considerably more frequent compared to others like "Bipolar" or "Personal Disorder". This imbalance can potentially lead to biased model performance, where the classifier may favor majority classes while underperforming on minority classes. To address this issue, we propose the use of alternative loss functions specifically designed to mitigate the impact of class imbalance. These include the Focal Loss  \cite{16}, which down-weights easy examples and focuses learning on hard misclassified samples; the Dice Loss  \cite{17}, originally used in image segmentation but adapted effectively for classification tasks with imbalanced data; and the Tversky Loss  \cite{18}, a generalization of Dice Loss that provides more control over the balance between false positives and false negatives. These loss functions help the model learn more robustly across all classes by giving more importance to underrepresented categories. Also standard techniques such as undersampling and oversampling were explored to mitigate the imbalance. While undersampling reduces the majority class to balance with the minority, it can lead to the loss of valuable information. Oversampling, on the other hand, duplicates or synthetically generates minority samples, which may cause overfitting or introduce unnatural patterns into the training process.

These resampling strategies were tested during the experimentation phase involving classical machine learning models to observe their effect on performance. However, due to the limited flexibility and potential drawbacks of these methods, they were not applied during the training of deep learning and transformer-based architectures.

Instead of altering the data distribution, the deep learning approach addressed the imbalance through the design of custom loss functions. These loss functions inherently increased the penalization for misclassifying minority classes, allowing the model to focus more on underrepresented samples without modifying the dataset itself.

During the data splitting process, the proportional class distribution was carefully preserved across both training and test sets to ensure a fair and representative evaluation.

Additionally, k-fold cross-validation was employed to enhance the robustness of the results. This method helped reduce the variance between training iterations and provided a more reliable assessment of model generalization, especially under the constraints of imbalanced data.

Another characteristic of our dataset is the variation in text lengths. While the dataset contains samples with a wide range of word counts, the majority of the entries tend to be relatively short in length. This observation suggests that most users express their mental states using brief sentences or short conversational snippets, which is a typical trait in real-world chat or social media-based mental health data. 

Figure 2. illustrates the distribution of text lengths in the dataset, highlighting the higher frequency of shorter samples.

\vspace{0.3cm}
\begin{figure}[ht]
\centering
\includegraphics[width=0.5\textwidth]{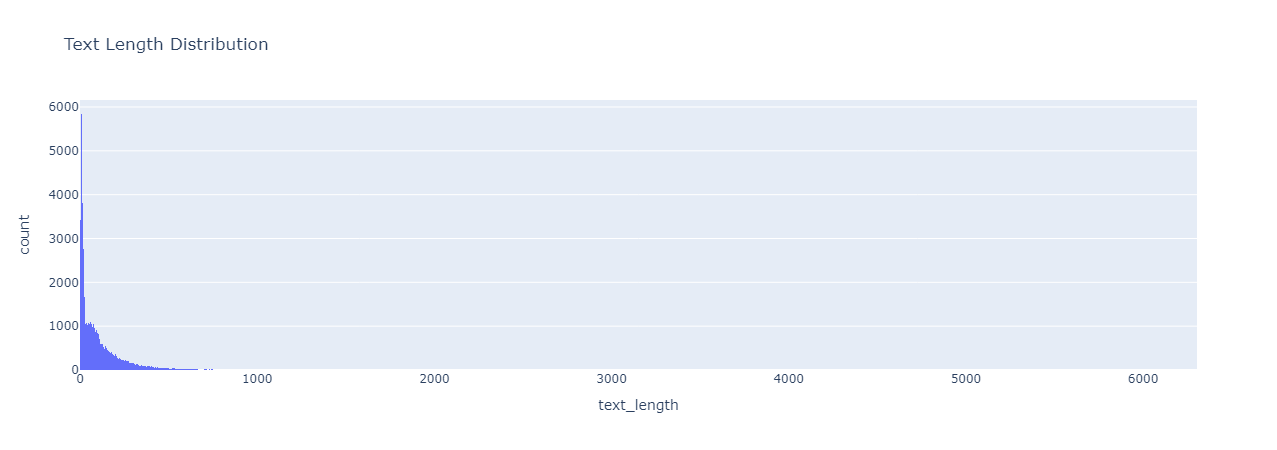}
\caption{Text Length distribution of the final dataset.}
\label{fig:class_distribution}
\end{figure}

The combined and refined dataset, along with all the corresponding data, embeddings and TF-IDF vectors, are publicly available and the links for these sources can be found on our GitHub repository.

\subsection*{Preprocessing}

To prepare the dataset for further analysis and model training, several preprocessing steps were applied. Initially, all text data were converted to lowercase to ensure uniformity. Unnecessary elements such as square brackets, hyperlinks, HTML tags, special tags, punctuation marks, and newline characters were removed to clean the raw text. Additionally, null values and redundant whitespaces were eliminated to refine the dataset.

Following this, a data augmentation process was applied to enhance the dataset without altering the semantic meaning of the original samples. Each text sample was first translated into French using TextBlob’s translator and then translated back into English. This back-translation technique helped generate semantically similar yet syntactically different sentences in the same language, enriching the diversity of the dataset.

In parallel with text normalization and augmentation, class imbalance in the dataset was also taken into account during the preprocessing phase. Given the uneven distribution of mental health categories, care was taken not to introduce bias during text generation or cleaning.

For classical machine learning models, preprocessing was followed by additional steps such as undersampling and oversampling. These techniques aimed to balance class distributions prior to training, allowing baseline models to perform more reliably on minority classes.

Although data augmentation via back-translation helped increase the diversity of training samples, further synthetic expansion methods were deliberately avoided. This decision was made to preserve the linguistic authenticity of the dataset and to prevent oversaturation with artificial text patterns, especially in more nuanced categories.

Instead of relying heavily on synthetic data manipulation, deeper architectural stages addressed the class imbalance through advanced training strategies, particularly via custom loss functions. These allowed the models to handle underrepresented samples more effectively, building on the cleaned and augmented dataset without further altering its natural structure.

After augmentation, the resulting text samples were re-cleaned through the same preprocessing pipeline to ensure consistency. The final version of the dataset was thus prepared for tokenization and vectorization processes, enabling effective training of downstream models.

While the aforementioned preprocessing techniques were applied universally across all models, the subsequent tokenization and vectorization steps varied depending on the model architecture. For traditional machine learning models, TF-IDF vector representations were utilized to train the models. For deep learning models that rely on a CNN-based backbone, pre-trained \texttt{word2vec-google-news-300} embeddings were employed to represent the textual data \cite{mikolov2013efficient}. 

For transformer-based models, including BERT \cite{devlin2019bert}, ALBERT \cite{lan2019albert}, DistilBERT \cite{sanh2019distilbert}, and RoBERTa \cite{liu2019roberta}, fine-tuning was performed using the respective pre-trained tokenizers and model-specific architectures. These models have demonstrated strong performance in various NLP tasks and were selected for their balance between computational efficiency and accuracy.

\section{Method}

In this section, we describe the methodology used to transfer knowledge from the teacher model to the student model, as well as the architecture and loss functions employed in the DualLatentGNet framework. Our approach leverages a teacher-student knowledge transfer mechanism, where the teacher model, trained with a CNN backbone and two decoders, provides guidance to the student model. This guidance is based on the teacher’s feature vectors, which are passed through a Gaussian Mixture Model (GMM) to create a probability distribution. The student model is trained to match the teacher’s feature distribution by minimizing the Euclidean distance and the difference in predicted Gaussian components. Additionally, we introduce a novel loss term, \(\mathcal{L}_{\text{latentG}}\), which combines the probability density function and Euclidean distance to enhance the learning process. Furthermore, we consider alternative loss functions such as Focal Loss, Tversky Loss, and Dice Loss to address class imbalance in the dataset, providing a more robust training procedure.

\begin{algorithm}[ht]
\SetAlgoLined
\KwIn{Teacher feature vectors: \texttt{teacher\_latent\_v}, \texttt{teacher\_pred\_logits}; dataset $(X, y)$}
\KwOut{Student model predictions and stored PDFs and Euclidean distances.}

\textbf{} \texttt{teacher\_feature\_vectors} $\gets$ concat[\texttt{teacher\_latent\_v}, \texttt{teacher\_pred\_logits}]\;
\textbf{} \texttt{gmm} $\gets$ FitGaussianMixtureModel(\texttt{teacher\_feature\_vectors})\;
\textbf{} Initialize the Student Model\;

\For{each $(X, y)$ in dataset}{
    \texttt{student\_feature\_vector} $\gets$ StudentModel($X$)\;
    
    \texttt{dist} $\gets$ EuclideanDistance(\texttt{student\_feature\_vector}, \texttt{teacher\_feature\_vectors})\;
    
    \texttt{pred\_pdf} $\gets$ \texttt{gmm.predict}(\texttt{student\_feature\_vector})\;
    
    \texttt{pdf} $\gets$ MostLikelyGaussian(\texttt{pred\_pdf})\;
    
    Store \texttt{pdf}, \texttt{dist}\;
}
\caption{Teacher-Student Knowledge Transfer Algorithm}
\end{algorithm}

\subsection*{Algorithm Explanation}

This algorithm describes a method for transferring knowledge from a teacher model to a student model. It consists of several key steps outlined below:

1. Teacher Feature Vectors: The teacher model’s feature vectors, consisting of the latent vectors and prediction logits, are concatenated into a single vector, `$teacher\_feature\_vectors$`.

2. Gaussian Mixture Model (GMM): A Gaussian Mixture Model (GMM) is fitted to the teacher’s feature vectors. The GMM helps model the distribution of the teacher’s feature vectors and will be used to guide the student model.

3. Student Model Initialization: The student model is initialized, and it will learn from the teacher model using the GMM and feature vectors.

4. Feature Vector Calculation: For each sample $(X, y)$ in the dataset, the student model calculates its feature vector, `$student\_feature\_vector$`, based on the input $X$. This represents the student’s understanding of the data.

5. Euclidean Distance Calculation: The Euclidean distance between the student’s feature vector and the teacher’s feature vectors is computed. This distance measures how similar the student’s features are to the teacher’s.

6. Gaussian Prediction: The GMM, trained on the teacher’s feature vectors, predicts which Gaussian distribution the student’s feature vector most likely belongs to.

7. Compute PDF of the Most Likely Gaussian: The probability density function (PDF) of the most likely Gaussian component is calculated. This value represents how closely the student’s feature vector matches the teacher’s knowledge distribution.

8. Store PDF and Euclidean Distance: The PDF and Euclidean distance are stored for each sample in the dataset. These values help guide the student model’s learning process, aligning it closer to the teacher’s knowledge.

This method allows the student model to leverage the teacher's knowledge by using a GMM to predict and align the student’s features with the teacher’s feature distribution. 

\subsection*{Units of the Architecture}
CNNBackbone is a sequential block of Conv-BN-ReLU-Conv-BN-Relu-MaxPool units. Latent vectors can be obtained by reshaping the output of the CNNBackbone and a simple linear layer to adjust the channels. Classification Decoder is a FC network to make predictions for classification task and Reconstruction Decoder is another FC network to reconstruct the input vector. A demonstration of the DualLatentGNet architecture can be found in Figure 3.

\begin{figure*}[ht]
    \centering
    \includegraphics[width=1.0\textwidth]{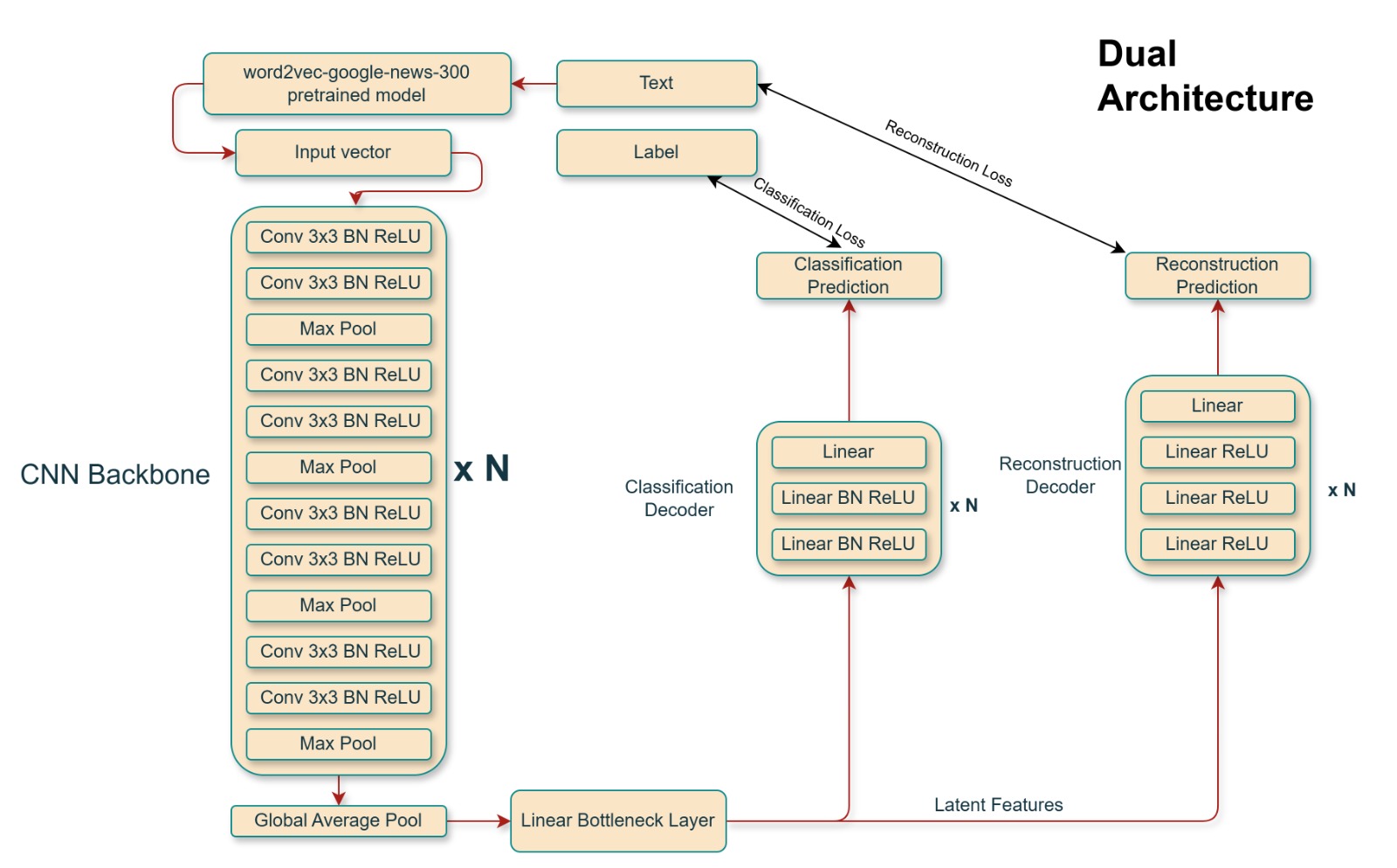}
    \caption{Dual Architecture Diagram}
    \label{fig:dual_architecture}
\end{figure*}

\subsection*{Intuition Behind the Method}

The core intuition behind the method is that individuals belonging to the same mental health class likely exhibit certain speech patterns or structures in their sentences, which share some underlying similarities. These similarities can be measured and learned by the teacher model.

We first train a teacher model that has a CNN backbone and two separate decoders for different outputs. One decoder operates like an autoencoder, trying to reconstruct the input, while the other decoder provides a classification output. The architecture called Dual Architecture[Figure 2]. This architecture ensures that the latent vectors produced at the end of the backbone are trained and fine-tuned by the backpropagation flow coming from both decoders. As a result, the model learns a meaningful latent vector representation of the input.

After training the teacher model, we obtain the latent vectors and classification prediction logits. These vectors are then concatenated into a single feature vector, denoted as $teacher\_feature\_vector$. This feature vector is used to represent the teacher's knowledge and its understanding of the input data.

Next, we fit a Gaussian Mixture Model (GMM) to these teacher feature vectors, assuming that the vectors come from multiple Gaussian distributions, one for each class in the dataset. Thus, we model the teacher’s feature vectors as originating from a number of distinct Gaussian components corresponding to the different mental health classes.

During the training of the student model, we use this fitted GMM to predict which Gaussian component each data point most likely belongs to. For each sample in every epoch, we calculate the Euclidean distance between the student’s predicted logits and the teacher’s predicted logits. Additionally, we compute the probability density function (PDF) value of the most likely Gaussian component for that sample.

Both the PDF and Euclidean distance values are stored and later used in the loss function. The student model’s goal is to minimize the difference between its predicted logits and the teacher’s, as well as to match the predicted Gaussian component with the true one. Specifically, we aim to minimize the terms $(1 - p)$ and $dist $ during training, where $p$ is the probability of the most likely Gaussian component, ensuring that the student model aligns with the teacher’s feature distribution and performs well in predicting the mental health class.

In this way, the student model progressively learns from the teacher model’s feature representations, helping it understand the structure of mental health data better, and ultimately leading to more accurate classification predictions.

After calculating the pdf and euclid distance, we can denote our latentG (latentGaussian) term as follows:

\[
\mathcal{L}_{\text{latentG}} = \alpha \left( 1 - p \right) + \beta \cdot \text{Dist}_{\text{euclid}}
\]

where \( \alpha \) and \( \beta \) are hyperparameters (scaling factors) that can be adjusted. The term \( (1 - p) \) represents the complement of the previously calculated PDF, meaning how much the sample is unlikely to come from the predicted Gaussian component. This is used to penalize the model when the prediction is far from the expected Gaussian component. The \( \text{Dist}_{\text{euclid}} \) represents the Euclidean distance between the predicted logits of the teacher and student models, quantifying the difference in their outputs.

So the Dual architecture's total loss will be:

\[
\mathcal{L}_{\text{total}} = \text{CE} \times \left(1 + \frac{e}{E} \times \mathcal{L}_{\text{latentG}}\right) + \text{MSE} \times \gamma
\]

where \( e \) is the current epoch, and \( E \) is the total number of epochs. The CE is used for classification (Cross Entropy Loss) and MSE (Mean Square Error Loss) for reconstruction. \( \gamma \) is a scaling factor that prevents one loss function from dominating the other when combining them.

In cases where the dataset suffers from class imbalance, the Cross Entropy Loss (CE) component can be replaced with alternative loss functions that are more suitable for such cases. These include:

\begin{itemize}
    \item \textbf{Focal Loss (FL)}:
    \begin{equation}
    \text{FL}(p_t) = -\alpha_t (1 - p_t)^\gamma \log(p_t)
    \end{equation}
    where $p_t$ is the model's estimated probability for the true class, $\alpha_t$ is a weighting factor, and $\gamma$ is the focusing parameter.
    
    \item \textbf{Tversky Loss (TL)}:
    \begin{equation}
    \text{TL} = 1 - \frac{TP}{TP + \alpha \cdot FP + \beta \cdot FN}
    \end{equation}
    where $TP$, $FP$, and $FN$ refer to the true positives, false positives, and false negatives respectively, with $\alpha$ and $\beta$ being tunable hyperparameters.

    \item \textbf{Dice Loss (DL)}:
    \begin{equation}
    \text{DL} = 1 - \frac{2 \cdot TP}{2 \cdot TP + FP + FN}
    \end{equation}
\end{itemize}

These loss functions aim to improve the learning process in scenarios with imbalanced class distributions by adjusting the influence of each training sample during backpropagation.

So the new LatentGLoss can be changed as below or their mixture.
\[
\mathcal{L}_{\text{new}} = \text{FL} \times \left(1 + \frac{e}{E} \times \mathcal{L}_{\text{latentG}}\right) + \text{MSE} \times \gamma
\]
 
\[
\mathcal{L}_{\text{new}} = \text{TL} \times \left(1 + \frac{e}{E} \times \mathcal{L}_{\text{latentG}}\right) + \text{MSE} \times \gamma
\]
\[
\mathcal{L}_{\text{new}} = \text{DL} \times \left(1 + \frac{e}{E} \times \mathcal{L}_{\text{latentG}}\right) + \text{MSE} \times \gamma
\]

\bigskip

\section{Experiments}

In this section, we present the experimental studies conducted throughout the development and evaluation of our proposed method. A wide range of machine learning, deep learning, and transformer-based models have been implemented, trained, and compared under various preprocessing and training setups. All the scripts and notebooks used in these experiments are publicly available on the GitHub repository. Also the training hyperparameters are available in Appendix section. 

\subsection{Machine Learning Models with TF-IDF Vectors}

After preprocessing and cleaning the dataset, each sample was vectorized using TF-IDF (Term Frequency–Inverse Document Frequency) representations. The following conventional machine learning algorithms were trained using these vectors: Logistic Regression, Naive Bayes, Support Vector Machines (SVM), Decision Tree, Random Forest, k-Nearest Neighbors (KNN), XGBoost, and Gradient Boosting.

A 3-fold cross-validation approach was adopted for all models, and the best hyperparameters were selected via grid search to ensure optimal performance. The top-performing models were further analyzed to identify the most informative features (i.e., words or expressions). Observed correlations between certain features and specific classes inspired the development of our proposed approach, which leverages latent representations aligned with such semantics.

\subsection{Initial Deep Learning Experiments with Word2Vec Embeddings}

In the initial deep learning experiments, word embeddings were generated using the pre-trained Word2Vec Google News 300-dimensional vectors~\cite{mikolov2013efficient}. The cleaned text dataset was tokenized, and each word was mapped to its corresponding embedding vector. These embeddings were then passed into a vanilla TextCNN architecture, which was trained for classification.

In parallel, a basic autoencoder architecture was trained on the same embeddings to learn latent vector representations. However, no clear correlation between the distribution of latent vectors and the classification outcomes was found. Furthermore, applying a Gaussian Mixture Model (GMM) over the latent space yielded insignificant results.

To further evaluate the performance of sequential models on mental health-related text data, additional experiments were conducted using Recurrent Neural Network (RNN) architectures. Specifically, simple RNN, Long Short-Term Memory (LSTM), and Gated Recurrent Unit (GRU) models were implemented. Instead of relying on pre-trained word embeddings, a learnable embedding layer was initialized and trained jointly with the models, allowing them to adaptively learn task-specific representations from scratch.

Each input sentence was tokenized and passed through the custom embedding layer, producing dense vector sequences which were then processed by the respective recurrent layers. These models captured sequential dependencies and context within the data, with LSTM and GRU outperforming the vanilla RNN due to their improved ability to model long-range dependencies via gating mechanisms.

Although these models showed enhanced understanding of temporal patterns in the text, their classification performance remained lower than that of the proposed dual-model architecture and transformer-based methods. Nonetheless, their results contributed to a broader understanding of model behavior across different architecture families.

\subsection{Dual Architecture and Latent Representation Alignment}

To address the limitations observed in the initial experiments, a Dual Architecture was proposed. This architecture combines a reconstruction decoder (for autoencoding) and a classification head, both contributing to the learning process of the latent vector through backpropagation. The dual nature of this architecture enabled the latent space to encode more meaningful and class-discriminative representations.

A teacher model was then trained based on this dual setup. To improve clustering and separation within the latent space, GMMs were fitted over combined feature vectors (concatenation of latent vectors and predicted logits). This strategy led to significantly more meaningful component assignments.

Subsequently, a new penalty function was developed based on the difference between the GMM probability density function (PDF) and the Euclidean distances of combined feature vectors. This formulation was incorporated into the loss function of a student model, which was trained under the guidance of the teacher model.

\subsection{Transformer-Based Models}

In addition to traditional ML and CNN-based deep learning architectures, transformer-based models such as BERT~\cite{devlin2018bert}, ALBERT~\cite{lan2019albert}, DistilBERT~\cite{sanh2019distilbert}, and RoBERTa~\cite{liu2019roberta} were fine-tuned on the dataset. For each model, the pre-defined tokenizer specific to that architecture was utilized to tokenize the input text, ensuring compatibility with the model's vocabulary and training dynamics.

\subsection{Comparative Framework}

Through these extensive experiments, our proposed architecture was compared against classical ML approaches, CNN and RNN-based deep learning methods, and state-of-the-art transformer models for text classification. The comparison results are reported and discussed in the subsequent \textbf{Results and Discussion} section.

\bigskip

\section{Results and Discussion}

In this section, we present and analyze the performance of all trained models. Both classical Machine Learning (ML) models and Deep Learning (DL) models were evaluated on the same test set to ensure comparability. The results demonstrate the strengths and weaknesses of each approach in terms of standard evaluation metrics, including Accuracy, Precision, Recall, and F1-Score.

For each model, different hyperparameters were tested with various training configurations, and the best-performing model results on the test dataset for each model type are presented in the Table I. The training scripts and Jupyter notebooks used for all models are available in the project's GitHub repository. Ready-to-use scripts for each model are provided for easy access.

\begin{table}[h]
\centering
\resizebox{\columnwidth}{!}{%
\begin{tabular}{lcccc}
\toprule
\textbf{Model} & \textbf{Accuracy} & \textbf{Precision} & \textbf{Recall} & \textbf{F1-Score} \\
\midrule
\hline
Logistic Regression & 86.80 & 87.00 & 87.00 & 87.00 \\
Naive Bayes         & 71.88 & 74.00 & 72.00 & 72.00 \\
SVM                 & 88.52 & 88.53 & 89.34 & 88.70 \\
Decision Tree       & 92.25 & 92.00 & 92.00 & 92.00 \\
Random Forest       & 93.41 & 94.00 & 93.00 & 93.00 \\
KNN                 & 39.35 & 59.00 & 39.00 & 30.00 \\
XGBoost             & 74.76 & 75.00 & 75.00 & 74.00 \\
Gradient Boosting   & 77.61 & 78.00 & 78.00 & 77.00 \\
\hline
RNN                 & 88.36 & 86.18 & 83.72 & 84.90 \\
LSTM                & 94.14 & 93.61 & 92.06 & 92.81 \\
GRU                 & 93.63 & 93.43 & 91.29 & 92.30 \\    
\hline
BERT (Finetuned)    & 94.83 & 94.81 & 94.83 & 94.80 \\
DistilBERT (Finetuned)    & 94.43 & 94.42 & 94.43 & 94.42 \\
AlBERT (Finetuned)    & 92.40 & 92.39 & 92.40 & 92.37 \\
RoBERTa (Finetuned)    & 93.38 & 93.37 & 93.38 & 93.37 \\
\hline
TextCNN             & 91.41 & 91.43 & 91.41 & 91.38 \\
DualTextCNN (Teacher) & 93.01 & 92.99 & 93.01 & 92.97 \\ 
\textbf{DualLatentGNet (Proposed)}  & \textbf{95.09} & \textbf{95.28} & \textbf{95.09} & \textbf{95.10} \\
\bottomrule
\end{tabular}%
}
\bigskip
\caption{Classification performance comparison of different models.}
\label{tab:results}
\end{table}

The results presented in Table \ref{tab:results} highlight the performance of various machine learning models and state-of-the-art transformer-based architectures on the classification task. While traditional machine learning models, such as Logistic Regression, Naive Bayes, and KNN, exhibit lower performance, advanced models like BERT, DistilBERT, and RoBERTa achieve competitive results, with BERT (Finetuned) achieving an impressive accuracy of 94.83\%. However, the proposed architecture, \textbf{DualLatentGNet}, outperforms all these models, including the transformer-based ones, with a top accuracy of \textbf{95.09\%}. This result demonstrates the effectiveness of our method in extracting rich and discriminative feature representations, even when compared to well-established models such as BERT and its variants.

Among the recurrent models, LSTM and GRU demonstrated strong performance, achieving 94.14\% and 93.63\% accuracy, respectively. These results confirm the effectiveness of gated recurrent architectures in capturing sequential patterns in text. LSTM slightly outperformed GRU, benefiting from its memory cell design that better handles long-term dependencies. In contrast, the vanilla RNN model lagged behind, with an F1-Score of 84.90\%, due to its known limitations in retaining long-term contextual information. Overall, these findings emphasize that even without pre-trained embeddings, learned representations through trainable embedding layers can yield competitive results when combined with well-designed sequential architectures.

However, the proposed architecture, \textbf{DualLatentGNet}, outperforms all these models, including the transformer-based ones, with a top accuracy of \textbf{95.09\%}. This result demonstrates the effectiveness of our method in extracting rich and discriminative feature representations, even when compared to well-established models such as BERT and its variants.

\begin{table}[h]
\centering
\begin{tabular}{l p{0.3cm} c c c c}
\toprule
\textbf{Model} & \textbf{Loss} & \textbf{Accuracy} & \textbf{Precision} & \textbf{Recall} & \textbf{F1-Score} \\
\midrule
\hline
DualLatentGNet              & \text{Dice}  & 94.30 & 94.43 & 94.41 & 94.42 \\
DualLatentGNet              & \text{Tversky}  & 94.56 & 94.75 & 94.22 & 94.49 \\ 
DualLatentGNet              & \text{CE}  & \textbf{95.09} & \textbf{95.28} & \textbf{95.09} & \textbf{95.10} \\
\hline
\bottomrule
\end{tabular}
\caption{Classification performance comparison of different loss functions.}

\label{tab:results}

\end{table}

For the proposed method, Tversky, and Dice Losses were tested as alternatives to Cross Entropy Loss, but the highest performance, as shown in the table, was achieved with Cross Entropy as in Table II. This can be attributed to the fact that the defined loss function may inherently address class imbalance. Specifically, by assigning different outputs to distinct Gaussian components and penalizing based on their corresponding probability density functions (PDFs), the model learns to make predictions based not on the count of classes but on the distribution in the feature space. This allows the model to focus on the spatial distribution of the data, potentially improving its generalization and accuracy.

The proposed architecture leverages a teacher-student knowledge transfer mechanism, where the teacher model is used to guide the student model's learning process. The teacher model, despite not being a transformer-based architecture, manages to produce feature vectors that effectively capture the underlying structure of the data. These feature vectors are then utilized to train the student model, significantly improving its classification performance. The success of \textbf{DualLatentGNet} can be attributed to its innovative approach of using a Gaussian Mixture Model (GMM) to model the distribution of teacher features, and incorporating this distribution into the student’s learning process.

Furthermore, \textbf{DualLatentGNet} surpasses traditional machine learning models, and more impressively, it competes with and even outperforms transformer-based models such as BERT and its variants. This highlights that the proposed approach not only excels in surpassing machine learning techniques but also demonstrates strong performance in comparison with state-of-the-art transformer models, offering a promising direction for future research in text classification tasks.

\section{Conclusion}

The significance of effective health classification cannot be overstated, particularly in the context of predicting mental states or other health-related conditions based on textual data. Accurately classifying text, such as sentences in patient reports or mental health assessments, or in any social platform, plays a vital role in improving the detection of various conditions and enhancing patient care. The preparation of datasets for such tasks is crucial, as it requires extensive preprocessing to clean, format, and normalize the data to make it suitable for training machine learning and deep learning models. In our study, we carefully curated and preprocessed the dataset, addressing issues such as missing values, tokenization, stopword removal, and normalization of text data, which are all essential steps in preparing health-related text data for classification tasks.

In this work, we proposed a novel approach for text classification tasks aimed at improving the performance of existing state-of-the-art models. The proposed architecture, DualLatentGNet, incorporates advanced techniques in feature extraction and model training, building upon both machine learning and deep learning paradigms, while introducing an innovative teacher-student framework. Our method, which focuses on leveraging knowledge transfer between models, has demonstrated substantial improvements in classification accuracy when compared to traditional machine learning models, as well as leading transformer-based architectures such as BERT.

Throughout the study, we evaluated a variety of models, including classical machine learning algorithms (Logistic Regression, Naive Bayes, SVM, Decision Trees, Random Forest, KNN, XGBoost, and Gradient Boosting), as well as more advanced transformer-based models or DL models (BERT, DistilBERT, AlBERT, RoBERTa, and TextCNN, RNN, LSTM, GRU). The results presented in Table I clearly show that DualLatentGNet outperforms not only the machine learning models but also transformer-based models, including BERT variants, achieving an accuracy of 95.09\%. This is a remarkable achievement, as transformer models such as BERT and its variants are typically seen as the leading approaches in modern NLP tasks. Our method challenges this paradigm, showing that transformer-based performance can be matched or even exceeded by a non-transformer architecture that leverages more efficient training methodologies and feature extraction processes.

The key contribution of this work lies in the development of the teacher-student model that utilizes a Gaussian Mixture Model (GMM) to enhance feature representations. The teacher model, although not based on transformer architecture, provides rich, informative feature vectors that help guide the student model’s learning. By effectively modeling the distribution of the teacher’s features, the student model is able to learn more discriminative and robust representations, leading to superior classification accuracy. This hybrid approach offers a significant improvement over traditional and transformer-based methods, demonstrating that competitive performance does not always require the complexity and computational cost associated with large transformer architectures.

Our experiments also show that the proposed DualLatentGNet architecture can rival transformer-based approaches, even outperforming them in some cases. This is particularly significant given that transformer models, due to their large size and computational demands, often require extensive hardware resources, especially in terms of GPU memory. In contrast, the proposed method, which utilizes a more compact architecture, offers competitive performance without the same level of computational expense. This is an important breakthrough, as it challenges the conventional wisdom that large transformer-based architectures are the only way to achieve state-of-the-art results. Our model’s efficiency and scalability present a promising alternative for real-world applications, where hardware limitations often pose a significant challenge.

Looking ahead, there are several areas where further improvements could be made. One of the key challenges we identified during this work is the issue of computational resources, particularly GPU memory limitations. While the DualLatentGNet method provides excellent results, scaling it to handle even larger datasets and more complex tasks does not necessarily require more GPU memory or larger models. This highlights the importance of developing more scalable approaches that can be easily adapted to a wide range of applications. Future work will likely focus on optimizing the network for even more efficient training and inference, potentially by incorporating methods such as model pruning, quantization, or distillation to further reduce computational requirements.

Moreover, while our method has proven to be competitive with transformer architectures, there is still room to explore how these approaches can be combined. We believe that a hybrid model that combines the strengths of both traditional models and transformer-based architectures could offer even better results. By continuing to explore these hybrid architectures, we can push the boundaries of what is possible in text classification and other natural language processing tasks.

In conclusion, the DualLatentGNet architecture presented in this work has demonstrated impressive results in text classification, outperforming traditional machine learning models and competing with leading transformer-based approaches. By focusing on efficiency, scalability, and feature extraction, our method provides an alternative to the highly resource-intensive transformer models, proving that performance and computational efficiency are not mutually exclusive. Future research should continue to explore more scalable methods that can handle the growing complexity of real-world tasks, while also ensuring that models remain practical and accessible for a wider range of applications.

\newpage
\begin{center}
\hspace*{\fill}
\bibliographystyle{IEEEtran}

\hspace*{\fill}
\end{center}

\newpage

\appendix
\section{Appendix}
Here you can find the hyperparameters that were selected for the best model trainings, which are shown in Table I and Table II. For further details on the machine learning model training parameters, please visit our GitHub page and refer to the relevant Jupyter Notebook.

\subsection{Model Training Hyperparameters}

\begin{table}[ht]
\centering
\begin{tabular}{|c|c|c|c|c|c|}
\hline
\textbf{Model} & \textbf{Epoch} & \textbf{Learning Rate} & \textbf{Batch Size} & \textbf{Weight Decay} & \textbf{Warmup Steps} \\
\hline
DualLatentGNet          & 500 & 0.01 & 32 & - & -   \\
\hline
TextCNN                 & 300 & 0.01 & 32 & - & -  \\
\hline
DualTextCNN (Teacher)   & 300 & 0.01 & 32 & - & -  \\
\hline
RNN                     & 200 & 0.001 & 64 & - & -  \\
\hline
GRU                     & 200 & 0.001 & 64 & - & -  \\
\hline
LSTM                    & 200 & 0.001 & 64 & - & -  \\
\hline
BERT (Finetuning)       & 10 & 3e-5 & 8 & 0.01   & 500  \\
\hline
DistilBERT (Finetuning) & 10 & 3e-5 & 8 & 0.01 & 500  \\
\hline
AlBERT (Finetuning)     & 10 & 3e-5 & 8 & 0.01 & 500  \\
\hline
RoBERTa (Finetuning)    & 10 & 3e-5 & 8 & 0.01 & 500  \\
\hline
\end{tabular}

\bigskip
\caption{Training Hyperparameters for Each Model}
\end{table}

\subsection{Additional Hyperparameters}

\begin{table}[ht]
\centering
\begin{tabular}{|c|c|c|c|c|c|}
\hline
\textbf{Model} & \textbf{Alpha} & \textbf{Beta} & \textbf{Gamma} & \textbf{Loss1} & \textbf{Loss2} \\
\hline
DualLatentGNet + CE & 0.56 & 0.44 & 75 & Cross Entropy & MSE \\
\hline
DualLatentGNet + TL & 0.56 & 0.44 & 75 & Tversky Loss & MSE \\
\hline
DualLatentGNet + DL & 0.56 & 0.44 & 75 & Dice Loss & MSE \\
\hline
\end{tabular}
\bigskip
\caption{Additional Hyperparameters for Proposed Model}
\end{table}

\end{document}